\documentclass[10pt,twocolumn,letterpaper]{article}

\usepackage{wacv}
\usepackage{times}
\usepackage{epsfig}
\usepackage{graphicx}
\usepackage{amsmath}
\usepackage{amssymb}
\usepackage{booktabs}
\usepackage{float}
\usepackage{adjustbox}

\usepackage{adjustbox}
\usepackage{caption}

\usepackage[accsupp]{axessibility} 

\usepackage{tabularx, booktabs}
\newcolumntype{?}{!{\vrule width 1.2pt}}
\newcolumntype{Y}{>{\centering\arraybackslash}X}

\def\wacvPaperID{250} 

\wacvalgorithmstrack   

\wacvfinalcopy 

\ifwacvfinal
\usepackage[breaklinks=true,bookmarks=false]{hyperref}
\else
\usepackage[pagebackref=true,breaklinks=true,colorlinks,bookmarks=false]{hyperref}
\fi

\begin{document}

\title{Audio-Visual Face Reenactment}
\author{Madhav Agarwal\\
IIIT, Hyderabad\\
\and
Rudrabha Mukhopadhyay\\
IIIT, Hyderabad\\
\and
Vinay Namboodiri\\
University of Bath\\
\and
C V Jawahar\\
IIIT, Hyderabad\\
\and
{{\tt\small\{madhav.agarwal,radrabha.m\}@research.iiit.ac.in, vpn22@bath.ac.uk, jawahar@iiit.ac.in}}
}


\twocolumn[{
\maketitle
\begin{center}
\includegraphics[width=\textwidth]{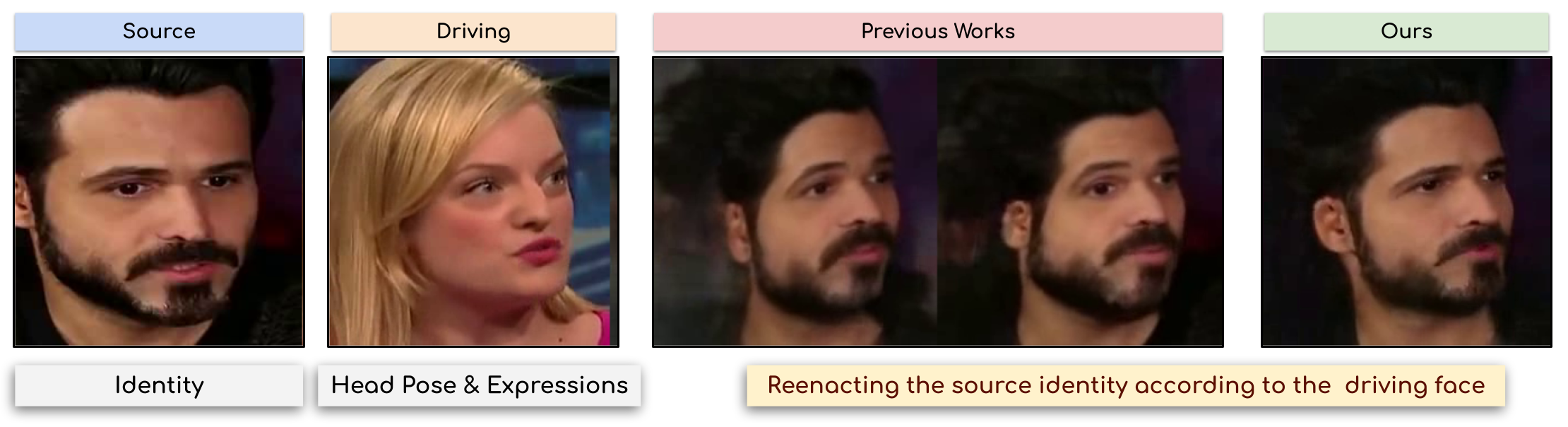}

\captionof{figure}{We propose AVFR-GAN, a novel method for face reenactment. Our network takes a source identity, a driving frame, and a small audio chunk associated with the driving frame to animate the source identity according to the driving frame. Our network generates highly realistic outputs compared to previous works like~\cite{Siarohin_2019_NeurIPS} and~\cite{siarohin2019animating}. Results from our network contain significantly fewer artifacts and handle things like mouth movements, eye movements, etc. in a better manner.}

\label{fig:teaser}
\end{center}
}]

\thispagestyle{empty}
\begin{abstract}
This work proposes a novel method to generate realistic talking head videos using audio and visual streams. We animate a source image by transferring head motion from a driving video using a dense motion field generated using learnable keypoints. We improve the quality of lip sync using audio as an additional input, helping the network to attend to the mouth region. We use additional priors using face segmentation and face mesh to improve the structure of the reconstructed faces. Finally, we improve the visual quality of the generations by incorporating a carefully designed identity-aware generator module. The identity-aware generator takes the source image and the warped motion features as input to generate a high-quality output with fine-grained details. Our method produces state-of-the-art results and generalizes well to unseen faces, languages, and voices. We comprehensively evaluate our approach using multiple metrics and outperforming the current techniques both qualitative and quantitatively. Our work opens up several applications, including enabling low bandwidth video calls. We release a demo video and additional information at \url{http://cvit.iiit.ac.in/research/projects/cvit-projects/avfr}.
\end{abstract}
\vspace{-20pt}

\section{Introduction}
Imagine your favorite celebrity giving daily news updates, motivating you to work out, or interacting with you on your mobile phone! What if a movie director could reenact an actor's image without actually recording the actor? Or, how about skilled content creators animating avatars in a metaverse to follow an actor's head movements and expressions in great detail? We can also reduce zoom fatigue~ \cite{zoom} by animating a well-dressed image of ourselves in a video call without transmitting a live video stream! These ideas seem fictitious, infeasible, and not scalable. But, how about animating or ``reenacting" a single image of any person according to a driving video of someone else? Face reenactment, thus, opens up many opportunities in a world that is becoming increasingly digital with each passing day.

Face Reenactment aims to animate a source image using a driving video's motion while preserving the source identity. Multiple publications have improved the quality of the generations. Existing works on talking head generation generally use a single modality, i.e., either visual\cite{hong2022depth,Siarohin_2019_NeurIPS,wang2021one,zakharov2019few} or audio features\cite{ji2021audio,wang2021audio2head,song2018talking}. Audio-driven talking head generation models are good at generating quality lip-sync; however, they have a serious drawback in handling non-verbal cues. The video-driven methods heavily rely on the disentanglement of motion from the appearance~\cite{lorenz2019unsupervised}. These methods generally use key points as an intermediate representation~\cite{Siarohin_2019_NeurIPS,hong2022depth,wang2021one} and try to align the detected key points of source and driving frames. These works learn key points in an unsupervised manner and fail to focus on specific regions of the face. This stems from inadequate priors regarding the face structure or the uttered speech. The final quality of the generations also suffers from using a basic CNN-based decoder that fails to capture the sharpness present in the source image and generates blurred output video. As a part of this work, we provide a detailed review of different approaches in Section~\ref{sec:related_works}.




In this paper, we analyze the shortcomings of the current works and add key modules to our network. We introduce \textbf{A}udio-\textbf{V}isual \textbf{F}ace \textbf{R}eenactment \textbf{GAN} (\textbf{AVFR-GAN}), a novel architecture that uses both audio and visual cues to generate highly realistic face reenactments. We start with providing additional priors about the structure of the face in the form of a face segmentation mask and face mesh. We also provide corresponding speech to our algorithm to help it attend to the mouth region and improve lip synchronization. Finally, our pipeline uses a novel identity-aware face generator to improve the final outputs. Our approach generates superior results compared to the current state-of-the-art works, as shown in Section~\ref{sec:eval}. We comprehensively evaluate our method against several baselines and report the quantitative performance based on multiple standard metrics. We also perform human evaluations to evaluate qualitative results in the same section. Our proposed method opens a host of applications, as discussed in Section~\ref{sec:applications}, including one in compressing video calls. Our work achieves more than $7\times$ improvement in visual quality when tested at the same compression levels using the recently released H.266~\cite{h266} codec.


Our contributions are summarized as follows:\\
1. We use additional priors in the form of face mesh and face segmentation mask to preserve the geometry of the face.\\
2. We utilize additional input in the form of audio to improve the generation quality of the mouth region. Audio also helps to preserve lip synchronization, enhancing the viewing experience.\\
3. We build a novel carefully-designed identity-aware face generator to generate high-quality talking head videos in contrast to the high levels of blur present in the previous works.


\section{Related Work}
\label{sec:related_works}

\begin{figure*}
\begin{center}
\includegraphics[width=\linewidth]{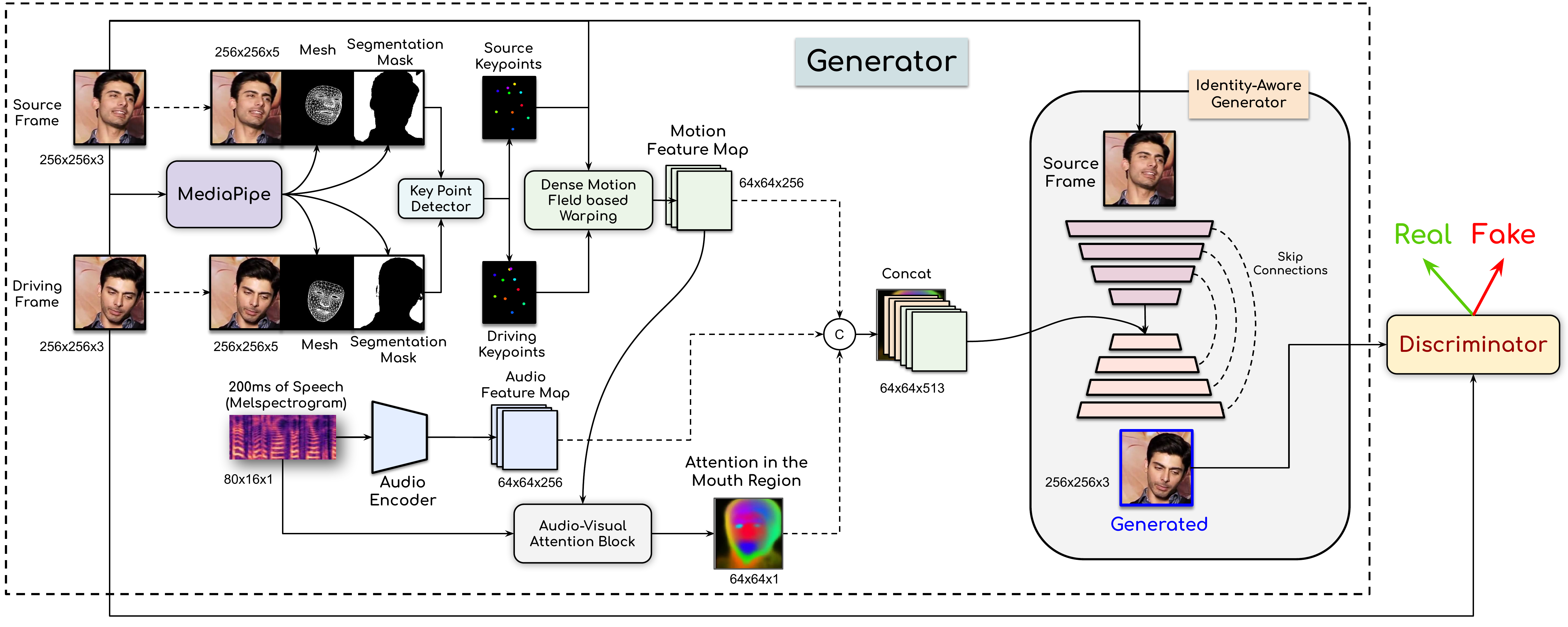}
\end{center}
\vspace{-20pt}
   \caption{The overall pipeline of our proposed Audio Visual Face Reenactment network (AVFR-GAN) is given in this Figure. We take the source and driving images, along with their face mesh and segmentation masks to extract keypoints. An audio encoder extracts features from driving audio and use them provide attention on lip region. The audio and visual feature maps are warped together and passed to the carefully designed Identity-Aware Generator along with extracted features of the source image to generate the final output.}
\label{fig:model_architecture}
\end{figure*}

Talking head generation works can be broadly classified in three categories based on the type of input they use to generate a talking head:
Text-driven~\cite{li2021write,tandon2021txt2vid,wang2011text}, Audio-driven~\cite{chung2017you,ji2021audio,lu2021live,song2018talking,wang2021audio2head,zhou2019talking,zhou2020makelttalk}, and Video-driven~\cite{hong2022depth,ren2021pirenderer,Siarohin_2019_NeurIPS,wang2021one,zhou2021pose} Talking Head Generation.

\paragraph{Text-driven Talking-head Generation}
Text-driven natural image generation~\cite{ramesh2021zero,reed2016generative} has recently seen a lot of progress in the computer vision community. Inspired by the recent success of GANs in generating static faces from text\cite{Wang_2021_WACV}, Li~\etal~\cite{li2021write} proposed a method to use text for driving animation parameters of the mouth, upper face and head. Txt2Vid~\cite{tandon2021txt2vid}  converts the spoken language and facial webcam data into text and transmits it to achieve low-bandwidth video conferencing using talking head generation. However, this method relies heavily on the generated speech, altering the original speaker's voice, prosody, and head movements in the video call. It depends on the quality of the Speech-to-Text module, which introduces grammatical errors and language dependency. Text as a medium has very little information about the head and lip movements; thus, we consider the problem ill-posed.

\paragraph{Audio-driven Talking-head Generation}
While text-driven methods suffer from a significant lack of adequate priors, we now move on to audio, a much more expressive and informative form of input. As the name suggests, audio-driven methods~\cite{chung2017you,ji2021audio,lu2021live,song2018talking,wang2021audio2head,zhou2019talking,zhou2020makelttalk} use only audio to animate a static face image. The first set of works like You-said-that?~\cite{chung2017you}, LipGAN~\cite{kr2019towards} and Wav2Lip~\cite{prajwal2020lip} achieved lip synchronization with given audio but failed to generate head movements in sync with the speech. These works used fully convolutional architectures and generated a single frame at a time without considering the temporal constraints. Eventually, a different class of works starting from Song~\etal~\cite{song2018talking} in 2018 and Zhou~\etal~\cite{zhou2019talking} in 2019, started using conditional Recurrent Neural Networks to model the temporal characteristics of a talking face. In 2020, Zhou~\etal~\cite{zhou2020makelttalk} published a landmark work that predicted dense flow from audio instead of directly generating the output video. The dense flow was then used to warp the source image to generate the final output. Several other well-known works like Emotional Video Portraits~\cite{ji2021audio} add an additional emotion label as input to create the talking head in the desired emotion. However, all of these works lack fine-grained control of the talking head and often contain a loopy head motion, and thus cannot be directly used in many applications.

\paragraph{Video-driven Talking-head Generation}
Finally, we move to video-driven methods, which use a driving video to get the motion and other facial features required to reenact a source image. Please note that the driving video and the source image may not have the same identity. Owing to the significant priors in driving video, the final generation quality of video-driven methods surpasses those of text-only and audio-only ones. 
The most influential work in this area, First-Order-Motion-Model (FOMM), was published by Siarohin~\etal~\cite{Siarohin_2019_NeurIPS} in 2019. The key idea was to estimate the motion field from sparse keypoints detected in both source and driving frames. The motion field was used to calculate dense flow and warp the source frame in a latent space. Several other works~\cite{wang2021one, hong2022depth} followed the same principle and added supplementary components to improve the quality. Face-vid2vid~\cite{wang2021one} used keypoint information in a 3D space, taking care of head rotation, among other things. DA-GAN\cite{hong2022depth} further added depth-aware attention to provide dense 3D facial geometry to guide the generation of motion fields. A similar approach in Motion-Representation-in-Articulated-Animation~\cite{siarohin2019animating} uses key regions instead of keypoints to generate the warpable motion field. Approaches like ICface\cite{Tripathy_2020_WACV} provide a method to control the pose and expressions of a face image using head pose angles and action unit values. Recently, Zhang~\etal~\cite{zhang2021flow} proposed using the three-dimensional morphable face model (3DMM) parameters to reenact a face image. They demonstrated that motion descriptor parameters for 3DMM can be derived from a driving video and, in turn, animate a static facial image. 

To the best of our knowledge, PC-AVS~\cite{zhou2021pose} is the only work that uses audio and video to formulate a low-dimension pose and motion code. Unlike FOMM, PC-AVS does not predict motion fields to calculate dense flow and warp the source image. Instead, they try to train their network to learn motion in a latent space inherently. While this allows them to achieve state-of-the-art lip sync, the generated video's overall quality is considered inferior to works like DA-GAN~\cite{hong2022depth}. In this work, we base our approach on FOMM's~\cite{Siarohin_2019_NeurIPS} principles and improve it with additional audio information. We also provide additional structural information to extract better geometries of the face. This allows us to use the best of both worlds and propose a novel network AVFR-GAN as described in the next section.

\section{Audio-Visual Face Reenactment GAN}

We present \textbf{A}udio-\textbf{V}isual \textbf{F}ace \textbf{R}eenactment GAN (\textbf{AVFR-Gan}), which takes a source image and a driving video plus audio to create high-quality talking head videos by preserving the source identity. As mentioned previously, we follow a similar strategy to that of FOMM~\cite{Siarohin_2019_NeurIPS} for our training pipeline. Instead of generating multiple frames in the form of a video, we handle the input in a frame-by-frame fashion. Our main goal is to estimate the motion between a source and a driving frame and then warp the source frame accordingly to generate an approximation of the driving frame.  Our model can be broadly divided into a Generator $M_{Gen}$ and a discriminator $M_{Disc}$ as shown in Figure~\ref{fig:model_architecture}. We first discuss the individual components present inside the generator. 

\begin{figure}
\begin{center}
\includegraphics[width=\linewidth]{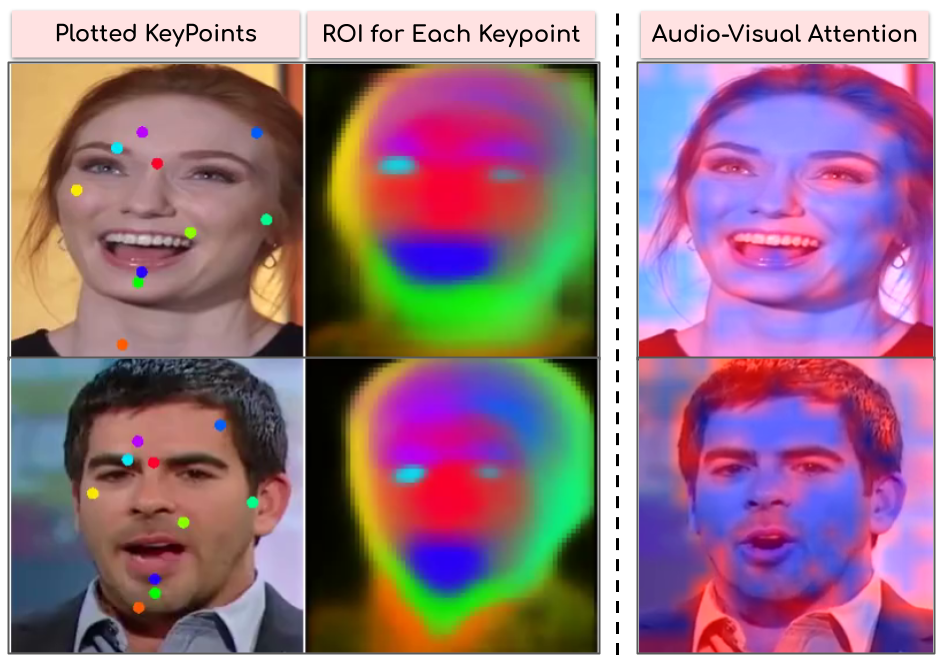}
\end{center}
    \vspace{-20pt}
   \caption{Illustration of keypoints detected (left), colour coded heatmap corresponding to each keypoint (centre) and the attention generated by our Audio-Visual Module (right). The ROI image shows that there are keypoints specific to the eye and mouth region. Attention image shows the important facial regions on which AVFR-Gan focuses.}
\label{fig:keypoint_attention}
\end{figure}

\paragraph{Additional Structural Priors to the Keypoint Detector}
We start with selecting a source frame $F_s$ and a driving frame $F_d$ both of dimensions $h \times w$. During training, both of these frames are selected from the same video. We pass these frames through mediapipe~\cite{lugaresi2019mediapipe} to generate a face mesh and a face segmentation map. We channel-wise concatenate the generated mesh and the segmentation mask with their respective images and create $5$ channel versions of the same. We term the concatenated source and driving frames as $I_s$ and $I_d$, respectively. We use these concatenated inputs to feed into our keypoint detector, $M_{kp}$. The addition of these priors helps us in providing the keypoint detector with more information about the respective structures of source and driving frames. Furthermore, the segmentation mask also provides the module with foreground and background information enabling the keypoints to be detected only from the foreground. We use the keypoint detector from FOMM~\cite{Siarohin_2019_NeurIPS} in our architecture. The keypoint detector $M_{kp}$ detects $K$ keypoints. More concretely, we can write,
\begin{equation}
\label{eq:1}
\{X_{T,n}\}_{n=1}^K = M_{kp}(I_T), T \in {s, d}
\end{equation}
The difference between the generated keypoints from the source and driving frames is used to calculate the motion field following FOMM. The motion field is then used to calculate dense flow and generate a warped feature map. We denote this feature map as Motion Feature Map, $Enc_{motion}$ as it captures the motion between the source and the driving frames. The dimension of this feature map is kept to be $\frac{h}{4} \times \frac{w}{4} \times c$. We plot sample keypoints detected in specific frames in Figure~\ref{fig:keypoint_attention} (left). Also, note that each keypoint has a specific region of interest in the generated motion field. We plot the heatmaps for each keypoint in Figure~\ref{fig:keypoint_attention} (middle). The heatmaps show that the regions of interest for each keypoints correspond to specific facial features. For example, the dark blue keypoint attends to the mouth region, green attends to the jaw, and sky blue attends specifically to the eye regions. Interestingly both of the eyes are attended by the same keypoint. 

\begin{figure}
\begin{center}
\includegraphics[width=\linewidth]{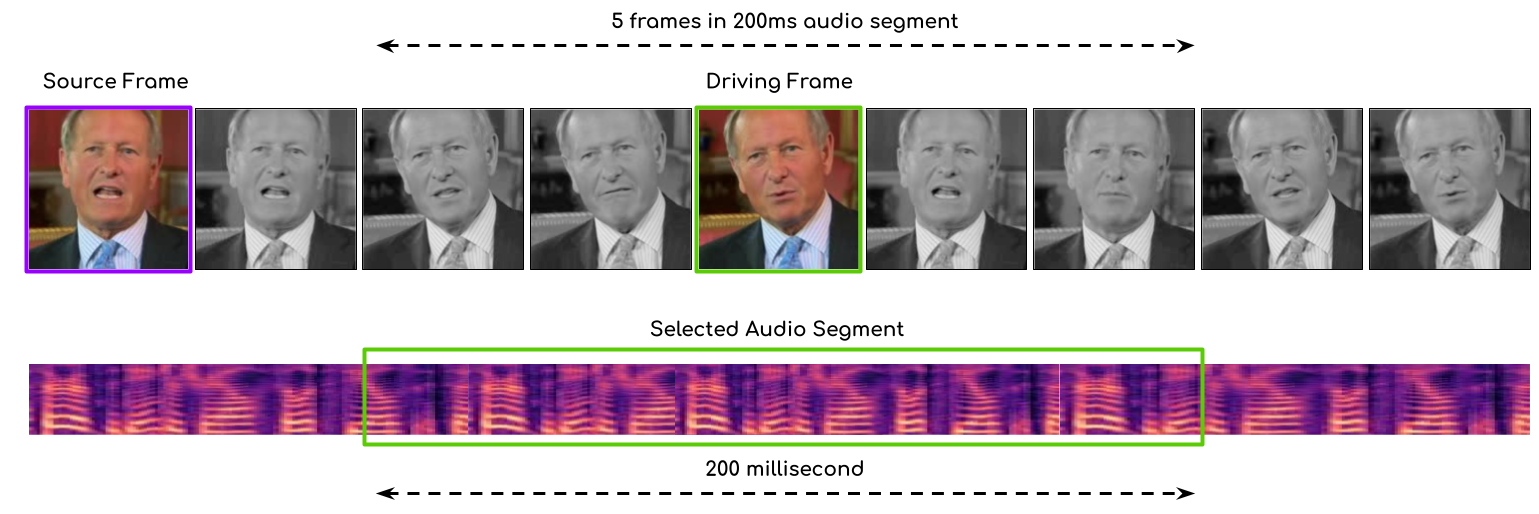}
\end{center}
\vspace{-20pt}
   \caption{Illustration of Audio window selector mechanism. It generates a 200ms spectogram such that the driving frame remains in the middle of the segment. In case of a 25 FPS video, a 200ms segment contains 5 frames.}
\label{fig:frame-select}
\end{figure}

\begin{figure}
\begin{center}
\includegraphics[width=\linewidth]{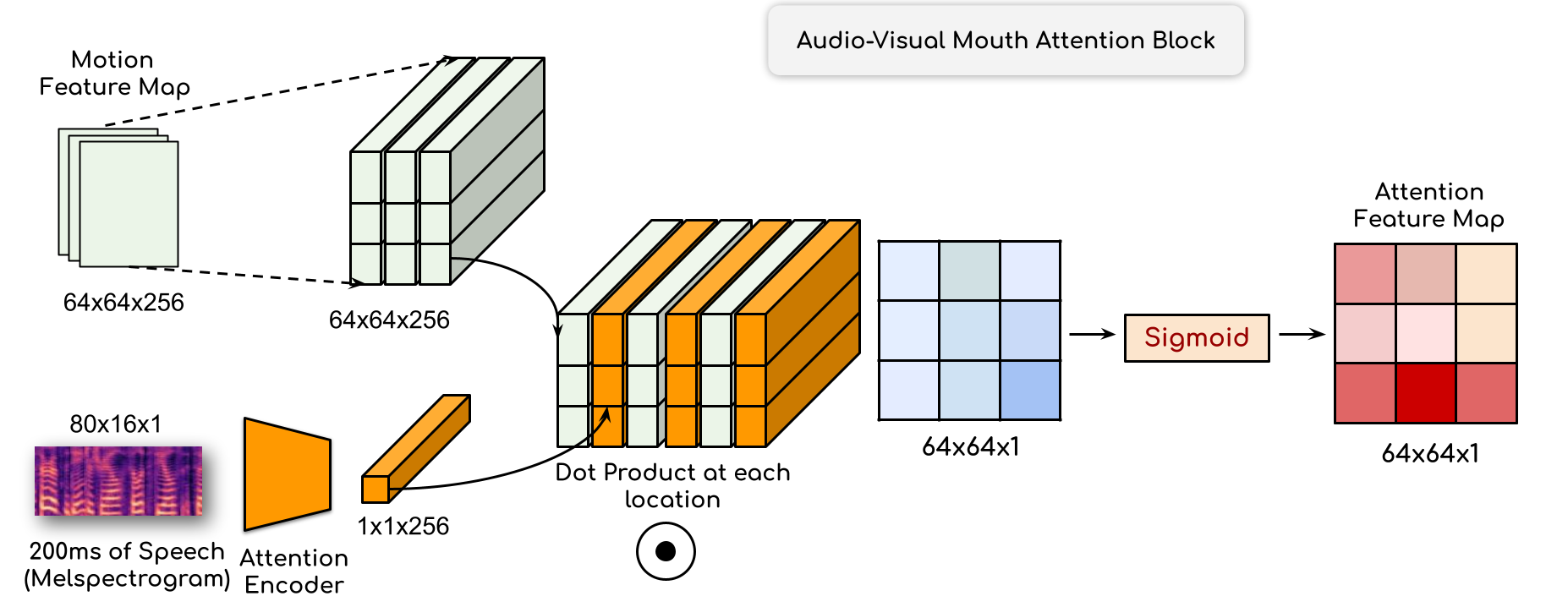}
\end{center}
\vspace{-20pt}
   \caption{Illustration of Audio Visual Attention module. Attention is generated by taking the dot product between a learned audio feature and visual features in each location, followed by a Sigmoid activation. }
\label{fig:audio-attention}
\end{figure}
\paragraph{Audio-conditioned Features}
Audio (mainly speech in our case) is an essential source of information that often accompanies a talking-head video. We decided to use the speech from the driving video to improve the quality of mouth movements in the generated video. While works like MakeItTalk~\cite{zhou2020makelttalk} have already generated head movements solely from audio, our goal is to only improve the mouth movements and transfer head motion directly from the driving video. Therefore, we follow the same strategy taken by lip-synchronization works like~\cite{chung2017you, kr2019towards, prajwal2020lip} to handle speech. We select the $200$ms window of speech around our driving frame $F_d$ such that $F_d$ is the middle frame in the sampling window. A graphical representation of the audio window selection is given in Figure~\ref{fig:frame-select}. We generate melspectrogram $I_{mel}$ from the speech window and feed it to a 2D CNN-based encoder. The audio encoder also outputs a feature map, $Enc_{aud}$, of $\frac{h}{4} \times \frac{w}{4} \times c$ dimension. We concatenate $(Enc_{motion}, Enc_{aud})$ along with the attention map generated as described next.   


\paragraph{Audio-Visual Attention}
Apart from improving the lip synchronization in the generated video, we propose using audio to specifically attend to the speaker's mouth region, enhancing the fine-grained details like teeth in the generated video. To do this, we pass $I_{mel}$ through an attention encoder generating an encoding $Enc_{query}$ of dimensions $1 \times 1 \times c$. We then take $Enc_{motion}$ of dimension $\frac{h}{4} \times \frac{w}{4} \times c$ and calculate the dot product at each location with $Enc_{query}$, generating a $\frac{h}{4} \times \frac{w}{4} \times 1$ matrix. We pass this through a Sigmoid layer to get the attention map $Enc_{attn}$ as shown in Figure~\ref{fig:audio-attention}. A formal definition of this block is given in Equation~\ref{eq:3}.

\begin{equation}
\label{eq:3}
\begin{gathered}
Enc_{attn} (i, j) = Sigmoid (Enc_{query} \odot Enc_{motion}(i, j)), \\ 
i \in \frac{w}{4}, j \in \frac{h}{4} 
\end{gathered}
\end{equation}

A visualization of the audio-visual attention can be found in Figure~\ref{fig:keypoint_attention}. As we can see, audio not only helps the model to attend to the mouth region but also helps the network attend to other regions like the eyes, which correlates to expressions from speech.

\paragraph{Identity-Aware Generator}
We propose a novel generator to decode the concatenated feature vector. We analyze the current decoders used in FOMM~\cite{Siarohin_2019_NeurIPS}, Face-Vid2Vid~\cite{wang2021one} and DA-GAN~\cite{hong2022depth}. We realize that the pipelines followed by the current works fail to  capture information from the source image directly. The network entirely depends on the warped features generated from the motion estimator to get the identity characteristics of the source speaker. Unfortunately, the warped features are forced to encode motion and fine-grained identity information, making it tougher to train. This ultimately causes the outputs to contain major artifacts and lose sharpness. We improve upon this and design an identity-aware face generator. We first concatenate $Enc_{motion}$, $Enc_{con}$ and $Enc_{attn}$ together to get the final warped features, generating $Enc_{dec}$. Instead of only feeding the warped features, we also feed in the source image $F_s$ separately to the UNet-shaped~\cite{unet} generator. The generator consists of an identity-encoder and a decoder. Both the encoder and decoder contain residual convolutional blocks inspired from Spatially Adaptive Normalization~\cite{park2019semantic}. The source image $F_s$ is first passed through an identity encoder to encode identity information. The output from the identity encoder is then concatenated with $Enc_{dec}$ and finally passed through the matching decoder with appropriate skip connections between the encoder and decoder blocks. The final output from the generator is denoted by $F_{gen}$. Our generator produces the sharpest output compared to the current state-of-the-art, as shown in the subsequent sections. 

\paragraph{Discriminator}
To improve the quality of our generated outputs, we also employ a standard discriminator, which is trained in a GAN setup along with the rest of the network. Our discriminator $M_{Disc}$, consists of a stack of Conv2D layers each followed by either spectral normalization~\cite{miyato2018spectral} or instance normalization~\cite{Ulyanov2016InstanceNT}. Each convolution block is followed by a Leaky ReLU activation~\cite{Maas2013RectifierNI}. The discriminator predicts a real or fake label and is trained to maximize the following loss function $L_{Disc}$ given in Equation~\ref{eq:4}.
\begin{equation}
\label{eq:4}
\begin{gathered}
\max_{M_{Disc}} L_{Disc} = \mathbb{E}_{x \sim p_{real}} \log M_{Disc}(x) + \\
\mathbb{E}_{F_{gen}} \log (1 - M_{Disc}(F_{gen}))
\end{gathered}    
\end{equation}

\paragraph{Losses used to train the Generator}

We use multiple loss functions similar to~\cite{Siarohin_2019_NeurIPS}. We use the $L1$ reconstruction loss between $F_d$ and $F_{gen}$. We also use the LPIPs~\cite{lpips} perceptual similarity loss (denoted by $L_{per}$) to improve the perceptual quality of the generated outputs. Finally, we employ the equivarience constraints $L_{eq}$ as described in the original FOMM paper. We refer the reader to ~\cite{Siarohin_2019_NeurIPS} for information regarding these constraints. While training the generator we also minimize the discriminator loss given in Equation~\ref{eq:4}. Therefore, we present our final loss function, Equation~\ref{eq:5}.
\begin{equation}
\label{eq:5}
\begin{gathered}
\min_{M_{Gen}} L_{Gen} = ||F_d - F_{gen}||_1 + \\ L_{per} + L_{eq} + \mathbb{E}_{F_{gen}} \log (1 - M_{Disc}(F_{gen}))
\end{gathered}    
\end{equation}

\paragraph{Inference Setting}
While we sample both $F_s$ and $F_d$ from the same video during training, our training strategy ensures that identity and motion information are well distilled. Therefore, our method allows for cross-identity face reenactment. During inference, we select a single image of a person as the source image $F_s$. Given a driving video of $N$ frames, $V_{i ... N}$, we pass each frame separately through our network along with $F_s$ and the corresponding audio segment of $V_i$ (denoted by $A_i$) to generate the final output as shown in Equation~\ref{eq:6}.
\begin{equation}
\label{eq:6}
\begin{gathered}
F_{Gen}^i = M_{Gen}(F_s, V_i, A_i), i \in 1 ... N
\end{gathered}    
\end{equation}

\begin{table*}[]
\centering
\begin{tabular}{|c|ccccccc||cc|}
\hline
 & \multicolumn{7}{c||}{Same-id Reenactment} & \multicolumn{2}{c|}{Cross-id Reenactment} \\ \cline{2-10} 
 &   \textbf{L1$\downarrow$} & \textbf{PSNR$\uparrow$} & \textbf{SSIM$\uparrow$} & \textbf{FID$\downarrow$} & \textbf{LMD$\downarrow$} & \textbf{AED$\downarrow$ }& \textbf{Sync$\uparrow$}      &    \textbf{FID$\downarrow$} & \textbf{Sync$\uparrow$} \\ \hline
 
 FOMM\cite{Siarohin_2019_NeurIPS} & 0.046 &  28.890 &  0.740 &  11.04    & 1.294 & 0.142 & 5.17  &  11.93  & 3.17 \\
 
 Face-vid2vid \cite{wang2021one} & 0.062  &  29.160 & 0.690  & 11.47 &  1.620 & 0.153 & 4.96 &  10.81 & 4.19 \\
 
 MRAA \cite{siarohin2019animating} & 0.040 &   23.351    &   0.64    &  11.36   & 1.280 & 0.135 & 3.10 & 15.61 & 3.96 \\
 
 PC-AVS \cite{zhou2021pose} &  0.081  &  23.750 & 0.620  & 14.32  & 1.843 &   0.180   & \textbf{6.76}   & 16.78 & \textbf{6.39}\\
 
 DA-GAN \cite{hong2022depth} & 0.036 &  31.220  & 0.804  &  9.10   & \textbf{1.278} & 0.129 & 5.01  & 9.40& 4.71\\
 
 AVFR-GAN (Ours) & \textbf{0.034}    &  \textbf{32.20} &  \textbf{0.824} & \textbf{8.48} &   1.280 & \textbf{0.127} & 5.45  & \textbf{9.05}  &  4.99\\ \hline
\end{tabular}
\caption{Comparison with state-of-the-art methods on Same-identity Reenactment and Cross-identity reenactment on VoxCeleb\cite{nagrani2017voxceleb} dataset. $\uparrow$ indicates larger is better, and $\downarrow$ indicates smaller is better.}
 \label{tab:sota_compression}

\end{table*}

\paragraph{Implementation Details}
In our experiments, we set $h=256, w=256$ and predict $K=10$ keypoints for training all our models. The model is trained using the Adam optimizer\cite{kingma2014adam} with a learning rate scheduler set at $60$ and $90$ epochs. The initial learning rate is set to be at $0.001$. The training time taken by model on $4$ NVIDIA RTX 3080Ti GPUs with a batch size of $10$ is around $10$ days. We train our model on the VoxCeleb~\cite{nagrani2017voxceleb} dataset, which contains 25 FPS videos. Thus, the $200$ms audio window consists of $5$ frames, of which the $3$rd frame is selected as the driving frame $F_d$. Any other random frame from the same video is selected as $F_s$ during training the network. More details about the network structure and other training characteristics are provided in the supplementary material.



\section{Experiments and Results}
\label{sec:eval}

We provide a comprehensive set of evaluations to measure the performance of our proposed method. We perform the quantitative assessment by following the standard benchmarks set by the previous works. We also perform extensive human evaluations to provide a qualitative assessment of the generated results. 

\paragraph{Evaluation Set}
We use the public test set of the VoxCeleb~\cite{nagrani2017voxceleb} dataset.
The dataset contains videos of celebrities. All the videos are preprocessed to $256 \times 256$. The test set contains 465 number of videos of different identities making up a total of 76 minutes. 
\vspace{-10pt}

\paragraph{Evaluation Metrics}
To provide an extensive evaluation of video reconstruction, we use several metrics to measure the performance of different works. We use the following metrics to measure various aspects of our generation. 
\textbf{L1}: It checks the average L1 distance between the generated and ground-truth video.
\textbf{LMD}: Landmark Distance calculates the distance between detected key points of ground-truth and developed video using a pre-trained facial landmark detector\cite{bulat2017far}. Please note that this metric was denoted by Average Keypoint Distance in~\cite{Siarohin_2019_NeurIPS}. However, we renamed it  Landmark Distance to avoid confusion with the keypoint detector module used in this work. 
\textbf{AED}: Average Euclidean Distance is used to evaluate the identity information. We use Openface\cite{amos2016openface} to find the feature vectors of generated and ground-truth video and then take the $L2$ distance between them.
\textbf{PSNR}: Peak Signal to Noise Ratio is used to evaluate the reconstruction quality of the generated image compared to the ground truth image.
\textbf{SSIM}: Structural Similarity Index evaluates the perceived changes in structural information of an image. 
We use it along with PSNR as it can also handle global illumination changes.
\textbf{FID}: Fréchet Inception Distance is used to compare the distribution of generated images with the ground truth image using the features extracted from an InceptionV3 model~\cite{szegedy2016rethinking}.
\textbf{Sync}: Syncnet confidence score is used to measure the amount of lip sync~\cite{chung2016lip}. 

\vspace{-10pt}

\paragraph{Comparison with State-of-the-Art Methods}

\begin{figure*}
\begin{center}
\includegraphics[width=\linewidth]{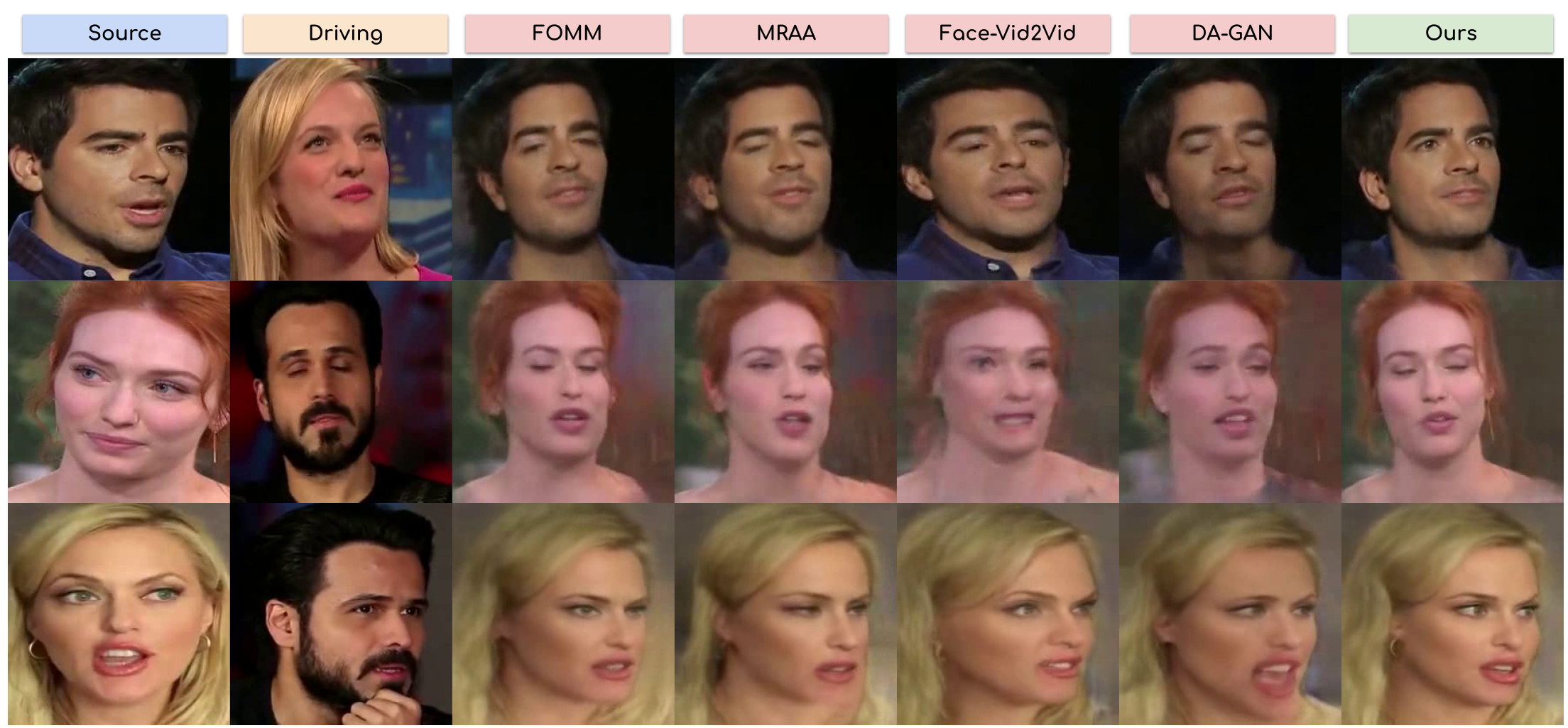}
\end{center}
\vspace{-20pt}
   \caption{Qualitative comparison on Cross-identity reenactment. Our method gives fewer artifacts, preserves facial structure and handle motion in a better way.}
\label{fig:cross_id}
\end{figure*}

We compare our work with the current methods published for the same task. To have a fair comparison, we use the official pre-trained models of FOMM~\cite{FOMM_github}, MRAA~\cite{MRAA_github}, PC-AVS~\cite{PCAVS_github} and DA-GAN~\cite{DaGAN_github} from their respective open-source implementations. For Face-Vid2Vid, we use an unofficial implementation in~\cite{facevid2vid_github}. All the pre-trained models and AVFR-GAN were trained on the same train split and evaluated on the test split of VoxCeleb\cite{nagrani2017voxceleb} using two inference strategies defined below. 

\vspace{-10pt}

\paragraph{Same-identity Reenactment} We perform the face reenactment task where the source frame and the driving video are of the same person. In this setting, we take the first frame of any video as the source frame and consider the rest of the video as the driving video. The audio chunks corresponding to each driving frame are also fed to the network as input. In this case, we expect the generated output to be as close to the original video as possible. We can therefore calculate metrics like L1, LMD, PSNR, and SSIM, which requires ground truth. We also calculate AED, FID, and Sync metrics for the generated outputs from all the models. From Table \ref{tab:sota_compression}, it is evident that our method outperforms all the other competing methods. The superior L1 and AED show that our model preserves identity information better. The improvement achieved by our model in terms of LMD indicates the improved structure of generated faces. Interestingly, our model generates improved eye movement in much more detail compared to the previous methods. We got state-of-the-art PSNR, SSIM, and FID scores, correlating with better visual quality. Finally, the sync quality achieved by our algorithm is superior to all the methods except PC-AVS, which performs slightly better in this metric.
\vspace{-10pt}

\paragraph{Cross-identity Reenactment} In this setting, we take a driving video for a different identity and animate a source image. The audio from the driving video is also given as input to the network, as usual. However, since the generated output does not mimic any specific ground truth, we use metrics that do not directly need the same. We use FID, which measures the distance between real and generated distributions and does not require one-to-one ground truths. We also use Sync to measure the quality of the lip sync in the generated video. As seen in Table~\ref{tab:sota_compression}, we achieve the best FID results and the second-best results in sync trailing only to PC-AVS.

\begin{figure*}
\begin{center}
\includegraphics[width=\linewidth]{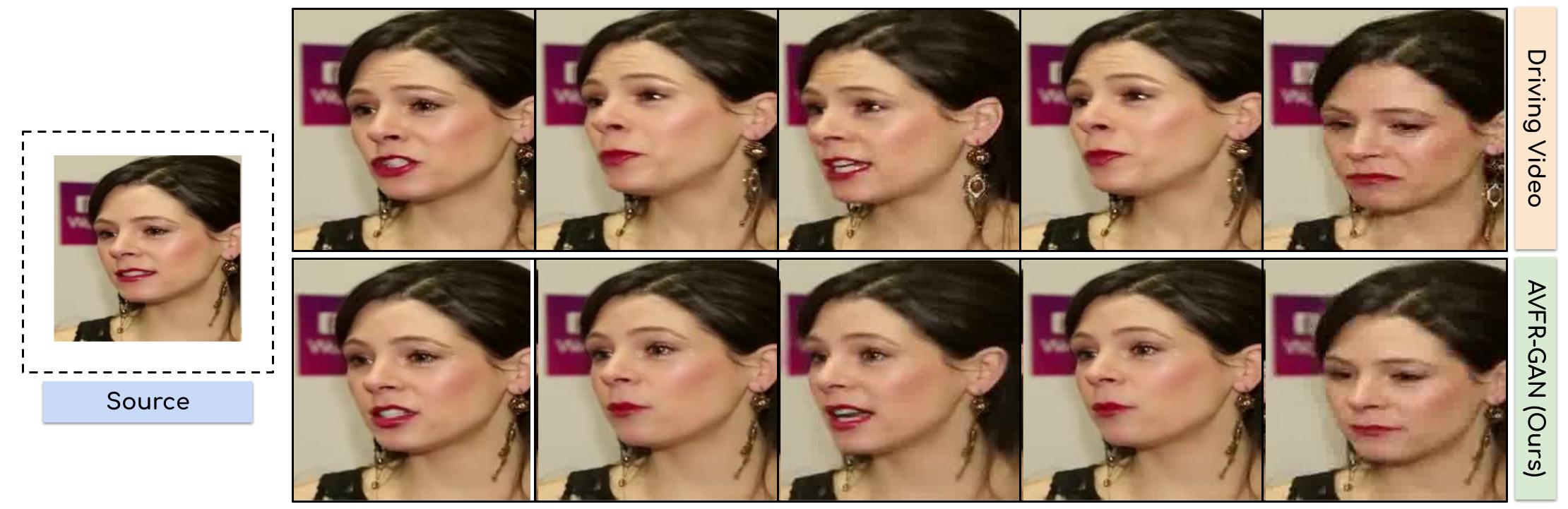}
\end{center}
\vspace{-20pt}
   \caption{Qualitative results on same-identity face reenactment. Upper row: Driving Video, Lower row: Generated Results}
\label{fig:same_id}
\end{figure*}

\paragraph{Human Evaluations}
Since our algorithm generates outputs directly meant for human consumption, we perform extensive human evaluations to ascertain the quality of the generations from our model from a human's perspective. We perform a study enrolling $20$ users. Each user is shown generated samples from the state-of-that-art method along with Ours. The users are also shown the source image and the driving video. We select $30$ samples from Cross-identity generations. Our user study shows corresponding results from each algorithm side by side, along with the source image and the driving video. The users are asked to rate each generated output based on three characteristics. The users rate the quality of 1. Head pose matching the driving videos, 2. Expressions matching the driving videos, 3. Identity preservation between the source image and the generated videos. The ratings are between $1$ to $5$, where $1$ corresponds to the worst and $5$ corresponds to the best. As seen in Table~\ref{tab:user_study}, our model consistently yields better results across all the criteria. Our model can enact a better head pose and match expressions of the driving video while preserving the source identities. 
\begin{table}[h]
\begin{center}
\begin{tabular}{| c | c c c |} 
 \hline
 \textbf{ } &  \textbf{HPMS$\uparrow$} &  \textbf{EMS$\uparrow$} &   \textbf{IPS$\uparrow$} \\ 
 \hline
 FOMM\cite{Siarohin_2019_NeurIPS} & 3.40 & 3.16 & 2.80\\
 Face-vid2vid \cite{wang2021one} & 3.70 & 3.12 & 2.66\\ 
 MRAA \cite{siarohin2019animating} & 3.26 & 3.06 & 2.50\\ 
 PC-AVS \cite{zhou2021pose} & 1.58 & 1.64 & 1.92 \\ 
 DA-GAN \cite{hong2022depth} & 3.98 & 3.82 & 3.10\\ 
 AVFR-GAN (Ours) & \textbf{4.56} & \textbf{4.22} & \textbf{3.94}\\ 
 \hline 

\end{tabular}
 \caption{User Study quantitative comparison. 'HPMS' represents Head Pose Matching Score, 'EMS' represents Expression Matching Score and 'IPS' represents Identity Preservation Score. $\uparrow$ shows higher is better. }
 \label{tab:user_study}
\end{center}
\end{table}


\section{Ablation Study}
Our proposed approach comprises addition of several key priors and the use of a better image generator. We check the contribution of each of these novel blocks in this section. For setting a baseline (very similar to FOMM), we remove Face Mesh, Face Segmentation, Audio Encoders, and used a basic CNN-based decoder architecture\cite{Siarohin_2019_NeurIPS,hong2022depth,wang2021one}. We add one module at a time to this baseline and train them on the same train-test split. We first add only face mesh and face segmentation to the baseline. We separately also check the effect of adding audio to the baseline. Finally, we combine the structural priors and audio to train a model without the novel identity-aware generator. We calculate SSIM, FID, and Sync metrics and report them in Table \ref{tab:ablation_study}.

\begin{table}[h]
\begin{center}
\begin{tabular}{| l | c c c |} 
 \hline
 \textbf{ } &  \textbf{SSIM$\uparrow$} &  \textbf{FID$\downarrow$} & \textbf{Sync$\uparrow$} \\ 
 \hline
 Baseline           & 0.74 & 11.04 & 5.17 \\
 + Structural Prior & 0.801 & 8.98 & 5.19\\ 
 + Audio Prior     & 0.79 & 8.69 & \textbf{ 5.48}\\ 
 + IAG    & 0.812 & 8.51 & 5.13\\ 
 AVFR-GAN  & \textbf{0.824} & \textbf{8.48} & 5.45\\
 \hline

\end{tabular}
 \caption{Ablation Study. The baseline represents the model without face mesh, segmentation, audio, and identity-aware decoder. '+ Structural Prior' represents Baseline with face segmentation and face mesh. '+ Audio Prior' represents Baseline with Audio encoders. '+ IAG' represents Baseline with Identity Aware Generator.  $\uparrow$ indicates larger is better, and $\downarrow$ indicates smaller is better.}
 \vspace{-20pt}
 \label{tab:ablation_study}
\end{center}
\end{table}
 As we observe clearly, the structural priors improve the SSIM significantly over baseline while audio improves the lip sync quality. We also observe that audio improves the visual quality (measured using FID) of the generations marginally. Finally, the identity-aware face generator gives a significant boost in terms of visual quality improvement.

\section{Applications}
\label{sec:applications}
Our work opens up several applications in the digital industry. Our method can revolutionize multiple industries. We can potentially replace recording famous celebrities in a studio environment costing thousands of dollars; we can animate a single picture of them based on home-recorded driving videos. Similar advances can also be made in the education sector, where online lectures are integral part of education. News readers can reduce their commute and present news from the comfort of their homes by animating their characters. We can also make video calls simpler in more than one way. We can replace the live video feed with a generated one reducing zoom fatigue. More importantly, this can lead to huge bandwidth reduction due to the compact keypoint-based representation, as already noted in~\cite{wang2021one}.  
\vspace{-20pt}
\paragraph{Low-bandwidth Video Conferencing}
Face reenactment methods can be easily extended for video compression. In the case of a video call between a sender and a receiver, we can first send a single high-resolution frame between the two and follow it up with sending keypoints detected by the keypoint detector for each frame. Our model can then generate the output frames at the receiver's end by considering the high-resolution frame as the source and keypoints from each of the driving frames, similar to the results shown in Figure~\ref{fig:same_id}. The $10$ keypoints each consist of $x$ and $y$ coordinates and four jacobians, all of which are represented as float values. Therefore, the total bits required to represent a $256 \times 256$ frame using FP16 representation is $10\times6\times16 = 960$ bits. Therefore, the Bits-per-Pixel(BPP) achieved by our model is $\frac{960}{256\times256} = 0.014$. We use the latest H.266 codec~\cite{h266} released in September of 2021 and compress the VoxCeleb test set at the same BPP. While the results generated by our algorithm achieve a FID of 8.48, the H.266 lags by a large margin at 58.32. This indicates the superior quality of the results generated using AVFR-GAN and provides a proof-of-concept for compressing video calls in future work.


\section{Further Discussions}
In this work, we propose a novel face reenactment network, Audio-Visual Face Reenactment GAN. Our network uses audio-visual cues to reenact a source image according to a driving video. We provide the network with additional structural priors and speech to improve lip synchronization. The final output quality also benefits from a novel identity-aware generator. The improvement in the quality of the generative networks has also led to concerns over its potential misuse. We, therefore, urge the users of any such works to use it ethically. We also encourage users to clearly mark the generated videos with a watermark. We believe these works will benefit and reduce manual effort in professional content creation.

{\small
\bibliographystyle{ieee_fullname}

\bibliography{egbib}
}

\end{document}